\documentclass[lettersize,journal]{IEEEtran}
\usepackage{amsmath,amsfonts}
\usepackage{algorithmic}
\usepackage{algorithm}
\usepackage{array}
\usepackage[caption=false,font=normalsize,labelfont=sf,textfont=sf]{subfig}
\usepackage{textcomp}
\usepackage{stfloats}
\usepackage{url}
\usepackage{verbatim}
\usepackage{graphicx}
\usepackage{cite}
\usepackage{amsmath}
\usepackage{mathtools}
\usepackage{multirow}
\usepackage{color, colortbl}
\definecolor{Gray}{gray}{0.9}

\usepackage{amssymb}
\makeatletter
\DeclareRobustCommand{\sqcdot}{\mathbin{\mathpalette\morphic@sqcdot\relax}}
\newcommand{\morphic@sqcdot}[2]{%
  \sbox\z@{$\m@th#1\centerdot$}%
  \ht\z@=.33333\ht\z@
  \vcenter{\box\z@}%
}
\newcommand{\ts}{\textsuperscript}

\newcommand*\sone{\vcenter{\hbox{\includegraphics[width=.8em]{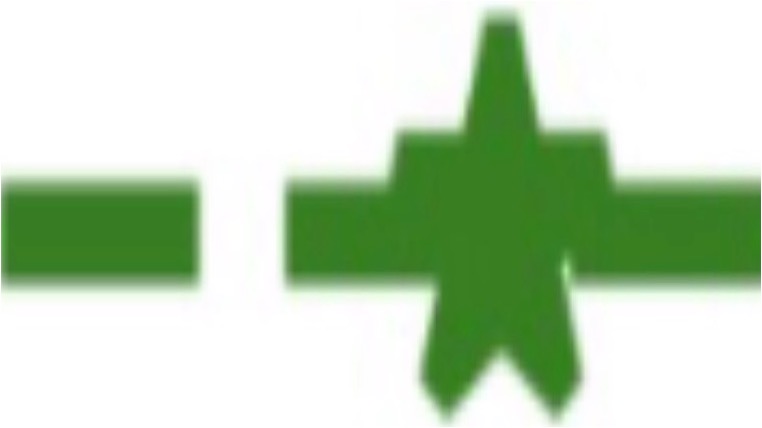}}}}
\newcommand*\stwo{\vcenter{\hbox{\includegraphics[width=.8em]{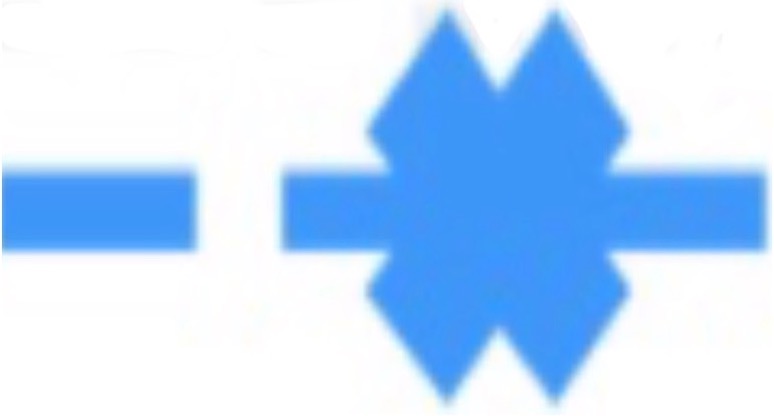}}}}
\newcommand*\sthree{\vcenter{\hbox{\includegraphics[width=.8em]{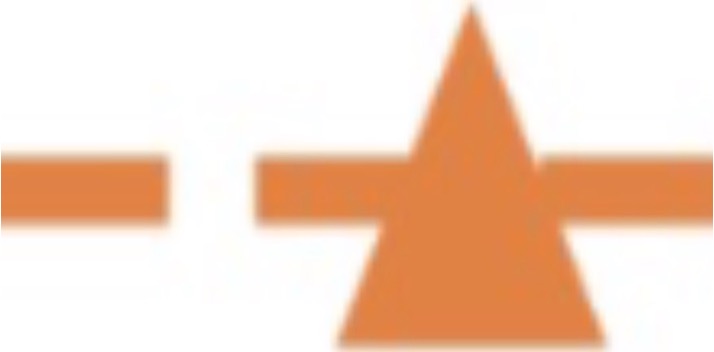}}}}
\newcommand*\sfour{\vcenter{\hbox{\includegraphics[width=.8em]{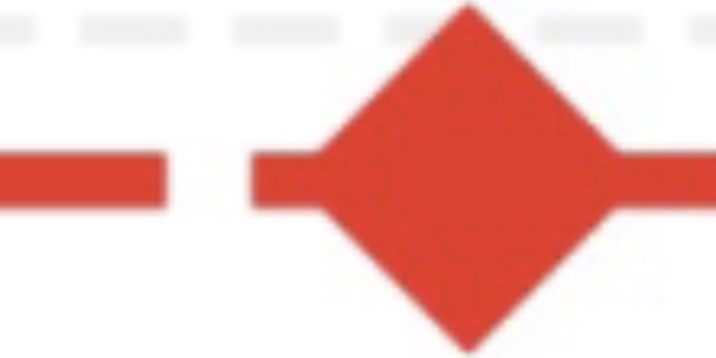}}}}

\usepackage{pifont}

\newcommand{\xmark}{\text{\ding{55}}}

\newcolumntype{C}{>{$\displaystyle}c<{$}}

\begin{document}

\title{Boosting Few-Shot Segmentation via Instance-Aware Data Augmentation and Local Consensus Guided Cross Attention}

\author{Li Guo *, 
    Haoming Liu *, 
Yuxuan Xia, Chengyu Zhang, Xiaochen Lu, and Zhenxing Niu

\thanks{*Equal contribution. Corresponding author: Li Guo.}
\thanks{Li Guo, Haoming Liu, Yuxuan Xia, Chengyu Zhang, and Xiaochen Lu are with New York University Shanghai, Shanghai 200126, China.}
\thanks{Zhenxing Niu is with Xi'an University of Electronic Technology, Xi'an 710126, China.}
}

\markboth{ }%
{Shell \MakeLowercase{\textit{et al.}}: A Sample Article Using IEEEtran.cls for IEEE Journals}

\IEEEpubid{}


\maketitle

\begin{abstract}
Few-shot segmentation aims to train a segmentation model that can fast adapt to a novel task for which only a few annotated images are provided. Most recent models have adopted a prototype-based paradigm for few-shot inference. These approaches may have limited generalization capacity beyond the standard 1- or 5-shot settings. In this paper, we closely examine and reevaluate the fine-tuning based learning scheme that fine-tunes the classification layer of a deep segmentation network pre-trained on diverse base classes.  To improve the generalizability of the classification layer optimized with sparsely annotated samples,  we introduce an instance-aware data augmentation (IDA) strategy that augments the support images based on the relative sizes of the target objects. The proposed IDA effectively increases the support set's diversity and promotes the distribution consistency between support and query images. 
   On the other hand, the large visual difference between query and support images may hinder  knowledge transfer and cripple the segmentation performance. To cope with this challenge, we introduce the local consensus guided cross attention (LCCA) to align the query feature with support features based on their dense correlation, further improving the model's generalizability to the query image. 
   The significant performance improvements on the standard few-shot segmentation benchmarks PASCAL-$5^i$ and COCO-$20^i$ verify the efficacy of our proposed method.
   
\end{abstract}


\section{Introduction}

\IEEEPARstart{S}{emantic} 
 segmentation has achieved tremendous success in recent years, thanks to the rapid development of deep learning algorithms. Despite the effectiveness of the deep learning models, they rely heavily on large amounts of annotated samples from well-established datasets. However, collecting sufficient dense annotated samples is both time-consuming and costly, especially for dense prediction tasks such as semantic segmentation and instance segmentation. To cope with this challenge, few-shot segmentation (FSS) aims to learn a generic segmentation model that can quickly adapt to novel classes in low-data regimes.   

 \begin{figure}[t]
\begin{center}
\includegraphics[width=0.95\linewidth]{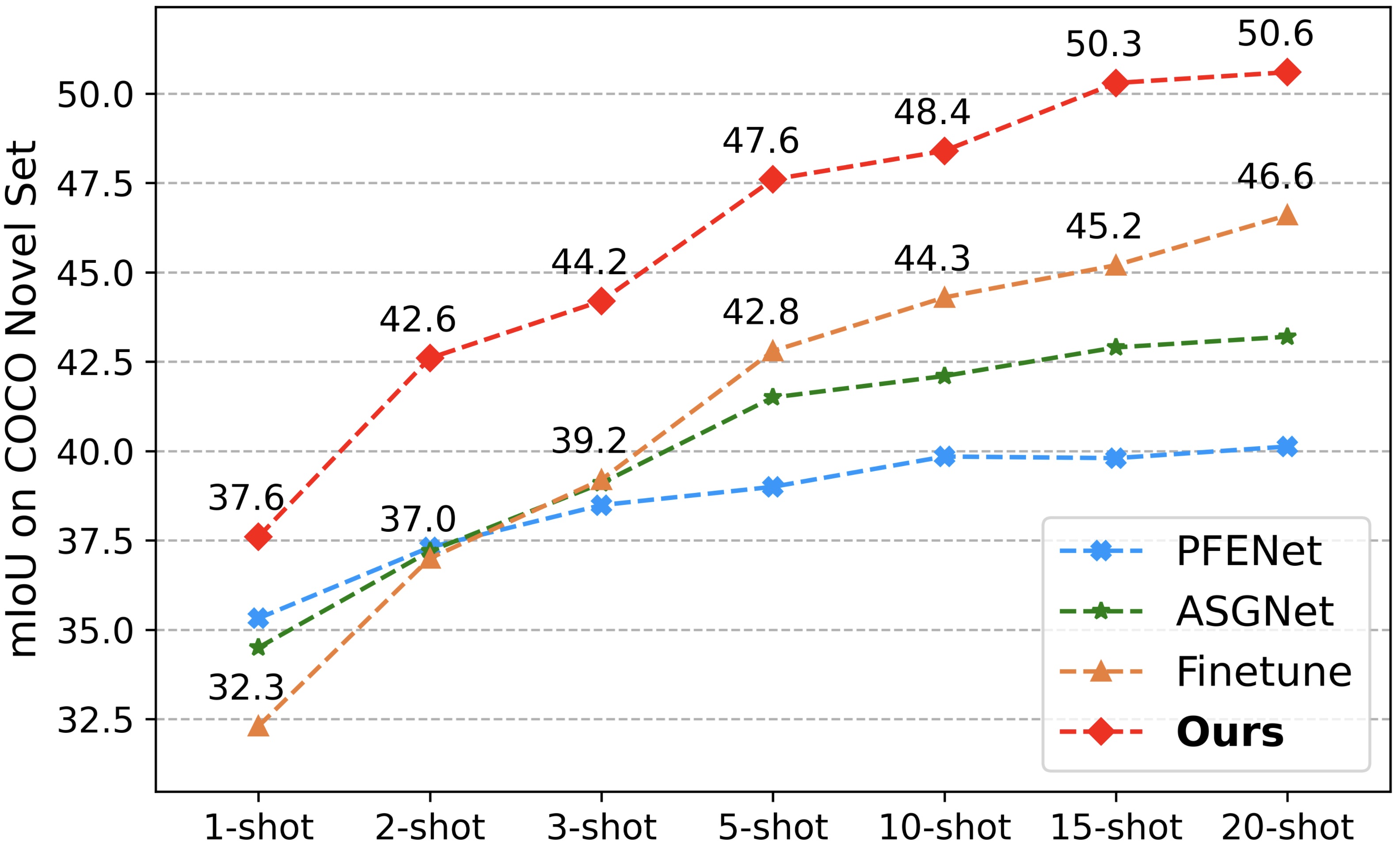}
\end{center}
   \caption{FSS performance (mIoU) on COCO-$20^i$ as the shot number increases. All models adopt a ResNet-50 backbone. ‘Finetune' represents the baseline fine-tuning approach described in Section \ref{ssec:overall}.}
\label{fig0}
\end{figure}

Even though the literature on FSS exhibits great diversity, they broadly fall into two categories. The first line of work adopts a prototype learning paradigm, which uses masked global average pooling \cite{zhang2019canet} or clustering \cite{li2021adaptive, yang2020prototype} to generate one or more prototypes and perform query object inference based on the dense comparison with the prototypes. On the other hand, several other works \cite{boudiaf2021few,lu2021simpler}  have adopted a two-stage fine-tuning based training strategy, which only fine-tunes the last layer, i.e., the classification layer of a segmentation model pre-trained on the data abundant base classes. Figure \ref{fig0} compares some popular few-shot segmentation algorithms under different few-shot settings. As is shown, the prototype-based approaches ($\stwo$ $\sone$) \cite{tian2020prior, li2021adaptive} achieve great performance in the extreme low-shot case, while their performance saturates quickly beyond the standard 1- or 5-shot settings. 
On the contrary, fine-tuning based learning paradigm ($\sthree$ $\sfour$) can sufficiently utilize the increasing number of support samples and reveals superiority in 5- or 10-shot cases, while it is significantly outperformed by its counterpart in the 1-shot case due to over-fitting. 


\IEEEpubidadjcol 

Compared to the fine-tuning-based methods \cite{boudiaf2021few,lu2021simpler}, which leverage the overall support set to optimize the classifier for separating pixels of different categories,  the prototype-based approaches \cite{tian2020prior, liu2020part, nguyen2019feature, wang2020few} perform correlation learning between the query and support images,  fully utilizing the feature similarity of each support-query pair.   We conjecture that  the direct correlation learning between the support-query image pair is the key ingredient that helps the prototype based approaches to excel at extreme low-shot settings.  It is, therefore, natural to ask whether it is possible to improve the fine-tuning based 
approaches by incorporating the direct correlation between query and support images. In this paper, we answer this question affirmatively.

In particular, we recognize that the classification layer fine-tuned on a few annotated samples inevitably overfits the support images. Therefore, instead of directly making inferences on the query object based on the query feature, we exploit the dense correlation between support and query image pairs to align the query feature with the support feature and perform classification based on the aligned query feature. However, the pixel-wise correlation between the two images can be very noisy \cite{rocco2018neighbourhood} due to the large visual difference, which undermines the reliability of the cross-attention module. Inspired by recent work on semantic correspondence \cite{rocco2018neighbourhood,  li2020correspondence, min2021convolutional, kim2022transformatcher}, we refine the dense correlation between the two images using local consensus constraints. Then we perform cross attention (Local Consensus guided Cross Attention) based on the enhanced correlation map to achieve feature alignment between the image pair. Incidentally, by integrating the local consensus guided cross attention module into the two-stage fine-tuning based training framework, our proposed method, dubbed Local Consensus guided Cross Attention Network (LC-CAN),  exhibits strong generalizability to the query images.

\begin{figure}[t]
\begin{center}
\includegraphics[width=0.8\linewidth]{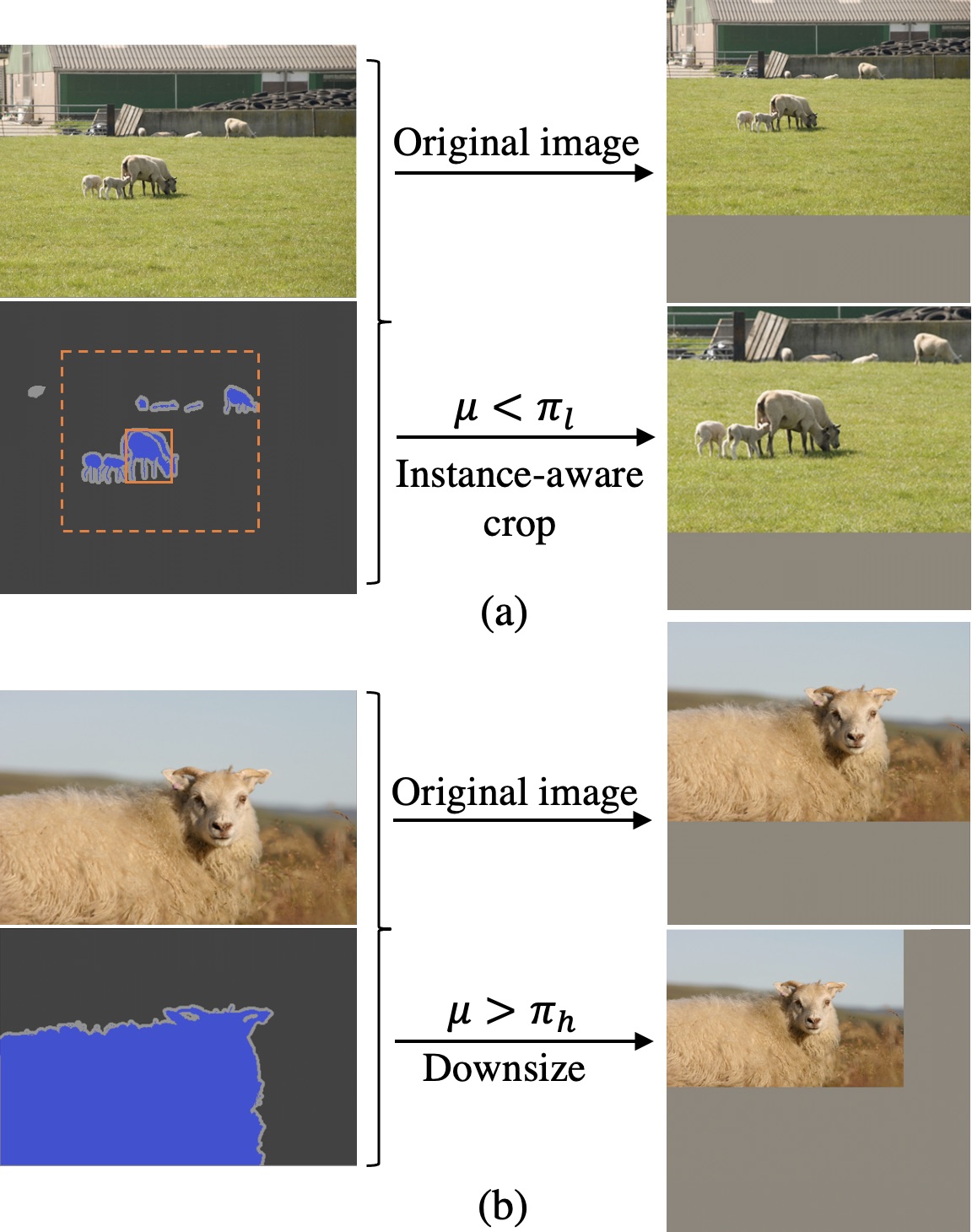}
\end{center}
   \caption{Visualization of instance-aware data augmentation. Original images are resized (while preserving the aspect ratio) and padded with grey pixels to the input image size. Given the foreground ratio $\mu$, we perform (a) instance-aware cropping when $\mu<\pi_l$ and (b) image downsizing when $\mu>\pi_h$ . The solid line in (a) is the bounding box of the largest target object, and the dashed line represents the cropping window.}
\label{fig1}
\end{figure}

In addition, we introduce a novel data augmentation mechanism to further alleviate the model over-fitting in fine-tuning based approaches. In ordinary supervised learning with sufficient labeled samples, the training data contains objects of different sizes and scales, whose distribution is relatively consistent with the testing data. Notably, under few-shot settings, the model is fine-tuned based on just one or few support images. Therefore the support set tends to 
form a biased representation of the ground truth distribution from which the test cases are sampled during evaluation. We propose Instance-aware Data Augmentation (IDA) to remedy this problem. As illustrated in Figure \ref{fig1}, IDA is implemented in an instance-aware manner: we first examine the relative sizes of the target objects in the support image, based on which an appropriate augmentation method is chosen to crop or downsize the image.   The key idea behind IDA is that we make the model exposed to support images of different scales while promoting distribution consistency between support and query images.

Overall, our contributions are summarized as follows:

\begin{itemize}
\item To the best of our knowledge, this is the first work to leverage data augmentation in fine-tuning based few-shot segmentation. The proposed instance-aware data augmentation strategy not only improves the diversity of support images and helps to promote distribution consistency between support and query images. 

\item We take advantage of the dense correlation between support and query images and design the local consensus guided cross attention to refine the query feature representation, further improving the model's generalizability to the query image. 

\item IDA augmentation strategy and the proposed LC-CAN framework cooperate in a synergistic fashion to further boost the model's segmentation performance on query images.  The improvements on standard few-shot segmentation benchmarks of PASCAL-$5^i$ and COCO-$20^i$ verify the efficacy of our proposed method.
\end{itemize}

\section{Related Work}

The related works are summarized into four aspects, including semantic segmentation, few-shot learning, few-shot semantic segmentation, and semantic correspondence. 

\subsection{Semantic Segmentation}
Semantic segmentation is a critical task in computer vision, and most existing approaches rely on fully convolutional networks with an encoder-decoder structure \cite{long2015fully}. However, semantic segmentation presents several challenges \cite{yuan2021ocnet}, such as reduced feature resolution that loses detailed spatial information and small receptive fields that fail to capture long-range dependencies. To address these issues, researchers have proposed various methods. For example, Yu {\it{et al.}} \cite{yu2015multi} used dilated convolutions to aggregate multi-scale contextual information, and Chen {\it{et al.}} \cite{chen2017deeplab} developed the DeepLab framework, which includes atrous convolution and atrous spatial pyramid pooling for multi-scale segmentation. Zhao et al. \cite{zhao2017pyramid} proposed PSPNet, which uses a pyramid pooling module to aggregate feature information from different regions, and DeepLabv3 \cite{chen2017rethinking} employs the ASPP module to aggregate multi-scale information using dilated convolutions with different dilation rates. Despite the significant success of current deep semantic segmentation models, they still heavily rely on large amounts of supervised information, which limits their practical application in scenarios with limited labeled data.

\subsection{Few-shot Classification} 
Few-shot learning (FSL) is a challenging research problem that aims to learn transferable knowledge that can be generalized to new novel classes with a few labeled samples. Existing approaches for FSL can be loosely organized into two families: meta-learning based approaches and fine-tuning based approaches.

Meta-learning based approaches can be further divided into optimization-based and metric-learning based methods. Optimization-based approaches \cite{finn2017model,ravi2017optimization,Jamal_2019_CVPR} aim to optimize the gradient descent procedure to enable fast adaptation of the model to new tasks. For example, MAML equips the model with a good initialization, allowing for fast learning on a new task with only a few gradient updates. TAML further enhances the model's generalizability by imposing an unbiased task-agnostic prior on the initial model to prevent over-performance on certain tasks.
Metric-learning based approaches \cite{koch2015siamese, vinyals2016matching, snell2017prototypical, ye2020few} learn a distance metric that compares the similarity between the support set and the query set. Matching Network \cite{vinyals2016matching} learns a metric space to compare the similarity of samples and predict their labels. Prototypical Network \cite{snell2017prototypical} learns a feature space in which class prototypes are easily computed as the mean embedding of the support set. DN4 \cite{li2019revisiting} is a nearest neighbor neural network that replaces the image-level feature with a local descriptor based image-to-class measure. 

Another line of work in few-shot learning adopts a fine-tuning based framework \cite{chen2019closer, dhillon2019baseline, laenen2021episodes, qi2018low}, which leverages pre-trained models and adapts them to new tasks. Typically, these methods pre-train a feature extractor with cross-entropy loss on the base classes and adapt only the classifier to the novel class at test time. For example, Qi \textit{et al.} \cite{qi2018low} utilize weight imprinting to initialize the final layer weights for novel classes during low-shot learning. Chen \textit{et al.} \cite{chen2019closer} demonstrate that a simple fine-tuning based framework can achieve results comparable to state-of-the-art meta-learning approaches, while Dhillon \textit{et al.} \cite{dhillon2019baseline} improve the performance of fine-tuning based approaches using transductive learning. Laenen \textit{et al.} \cite{laenen2021episodes} investigate the usefulness of episodic learning and demonstrate that meta-learning-based feature extractor can be detrimental to the model's generalizability.


\subsection{Few-shot Segmentation}
Few-shot segmentation (FSS) is a natural application of FSL in dense prediction tasks, and it has attracted considerable attention after the pioneering work OSLSM \cite{shaban2017one}. Most of the recent works in FSS adopt the prototype learning paradigm, which utilizes the prototype extracted from support samples to facilitate query object inference. Notably, CANet \cite{zhang2019canet} applies masked global average pooling on support images for prototype learning and conducts dense comparisons between the prototype and the query features. Instead of using a single prototype, ASGNet\cite{li2021adaptive} and PMMs \cite{yang2020prototype} cluster foreground pixels of the support image to multiple prototypes to account for the intra-class variation and provide more accurate guidance on query image inference. PFENet \cite{tian2020prior} generates a training-free prior mask based on high-level support and query features and combines this prior with low-level features to predict the final query mask with a novel feature enrichment module. Building upon PFENet, DCP \cite{lang2022dcp} adopts a novel self-reasoning scheme to derive a series of support-induced proxies for dense comparison of the query image. 
QCLNet \cite{zheng2022quaternion} explicitly explores  
internal latent interaction between query and support images
by leveraging operations defined by the established quaternion algebra. MFNet \cite{zhang2022mfnet} effectively fuses multi-scale query information and multi-class support information into one query-support embedding.
RPMG-FSS \cite{zhang2023rpmg} extended PFENet by generating the  prior mask based on the high-level features from multi-view support images and query images. 

On the other hand, several recent works \cite{boudiaf2021few,lu2021simpler} argue that the feature extractor of a deep segmentation model pre-trained on base classes is sufficiently generalizable to unseen classes. Instead of meta-training the feature extractor, they freeze the feature extractor pre-trained on base classes and focus on fine-tuning the classification layer for adaptation to new tasks. Our work adopts a similar fine-tuning based training scheme. In addition, we introduce a novel data augmentation strategy to alleviate the over-fitting problem when training the classifier in a low-data regime.

\subsection{Semantic Correspondence}
Semantic correspondence aims to find correspondences between semantically similar images under challenging degrees of variations \cite{balntas2017hpatches, ham2017proposal}. Recent approaches build the correlation map based upon feature representations extracted from convolutional neural networks pre-trained on image classification tasks. An emerging trend is to employ 4D convolutions \cite{li2020correspondence, min2021convolutional, rocco2018neighbourhood} on the dense correlation map to identify spatially consistent matches with the local match-to-match consensus constraint. In addition, some recent approaches \cite{min2019hyperpixel, min2020learning, zhao2021multi} for semantic correspondence show that combining features at different semantic levels can help to generate reliable features representation and further improve the matching accuracy. Inspired by existing works in semantic correspondence, we also exploit high dimensional convolution and multi-level features to construct and refine the affinity map between support and query features, which is utilized to guide the feature alignment between the image pair. 

\section{Problem Formulation}
We follow the standard setup and annotations in few-shot semantic segmentation \cite{shaban2017one}. Specifically, we are given two datasets $\mathcal{D}_{base}$ and $\mathcal{D}_{novel}$ with disjoint category sets $\mathcal{C}_{base}$ and $\mathcal{C}_{novel}$ respectively, where $\mathcal{C}_{base} \cap \mathcal{C}_{test} = \emptyset$.  The goal is to learn a segmentation model from the base dataset $\mathcal{D}_{base}$ with sufficient annotated samples so that the model can generalize well on new tasks sampled from the novel classes. Following the common episodic training protocol, we sample a series of episodes from $\mathcal{D}_{base}$ and $\mathcal{D}_{novel}$  to simulate the few-shot scenario. Particularly, under $K$-shot setting, each episode is composed of
: (1) a support set  $\mathcal{S} = \{(x^s_k, m^s_k)\}_{k=1}^{K}$, where $x^s_k$ and $m^s_k$ are the support image and its corresponding binary mask for a specific category; and (2) a query set $\mathcal{Q}=(x^q, m^q)$ where $x^q$ is the query image and $m^q$ is the ground truth binary mask. 
During training, the model is optimized based on the training episodes sampled from $\mathcal{D}_{base}$ to learn a mapping from $\mathcal{S}$ and $x^q$ to a prediction $m^q$. At inference, we evaluate the few-shot segmentation performance on test episodes sampled from $\mathcal{D}_{novel}$.


\section{Methodology}

In this section, we will begin by describing a baseline model which adopts the fine-tuning based training strategy in Section \ref{ssec:overall}. In addition, we will identify the critical issues that limit its generalizability. To address these limitations, we will elaborate on our proposed LC-CAN framework in detail.

\subsection{Two-Stage Training Strategy} \label{ssec:overall}

In general, a semantic segmentation model comprises a CNN encoder, a CNN decoder, and a simple classifier. Given an input image $x$, the CNN encoder $f_x=E_{\phi}(x) $ gradually reduces the feature map resolution and captures higher semantic information. The decoder module $z_x=D_{\psi}(f_x)$ aggregates multi-scale features and recovers the spatial information. Then the output embedding $z_x$ from the decoder module is directly passed to a pixel-wise classifier $\hat{m}=p_\theta(z_x)$ to separate pixels from different categories.

The key objective in meta-learning is to learn a transferable feature embedding network that generalizes to any new task. Several works \cite{dhillon2019baseline, laenen2021episodes} in few-shot classification showed that feature extractor pre-trained with standard cross entropy loss on base classes generates powerful embeddings for downstream tasks and often outperforms its counterpart trained with meta-learning paradigm. Following these findings,
The fine-tuning based training scheme involves a two-phase training procedure, as detailed below: 


\textbf{Model Pre-training}. In the first stage, we train the feature extraction network (i.e., the encoder and decoder) on the whole base dataset $\mathcal{D}_{base}$. Specifically, we use PSPNet \cite{zhao2017pyramid} as our backbone segmentation model, which is trained with standard cross entropy supervision. The training details are given in section \ref{ssec:implement}.

\textbf{New Task Adaptation}. In the second stage, we fine-tune the pre-trained segmentation model for adaptation to new tasks. Specifically, we keep the encoder and decoder backbone frozen and train the classifier only. 

With such a two-phase training strategy, the classification layer fine-tuned on a few annotated samples inevitably overfits the support set. To alleviate this problem, we introduce instance-aware data augmentation in Section \ref{ssec:ida}, and local consensus guided cross attention in Section \ref{ssec: lcca} for improving the model's generalizability to the query image.




\subsection{Instance-Aware Data Augmentation} \label{ssec:ida}

Data augmentation is a simple way to increase the number of training samples and alleviate model over-fitting. Random resize and crop are effective data augmentation methods widely used in regular segmentation tasks and can help improve the model's generalizability to test images of different scales.  

Appropriate data augmentations can be even more beneficial under the setting of few-shot segmentation since the distribution of the limited support images can be very biased. To alleviate the distribution bias of the support set, the proposed Instance-aware Data Augmentation (IDA) adaptively augments the support image based on the relative size of its target objects.

Given a support image, we first compute the proportion of its foreground area against the overall image size based on its ground truth mask, and we denote the foreground proportion as $\mu$. We compare the relative size of the target object  $\mu$ with pre-set hyperparameters $\pi_l$ and $\pi_h$ to adaptively determine the augmentation method applied to the support image. In our experiments, we set the thresholds $\pi_l$ and $\pi_h$ to $0.15$ and $0.3$, respectively. 

When the foreground object is relatively small,  i.e., $\mu < \pi_l$, we generate the augmented image using instance-aware crop as illustrated in Figure \ref{fig1}(a). When there are multiple foreground objects, we first identify the largest object which resides in the largest connected component of the foreground area. Then we crop the image to cover the largest foreground object. Particularly, for the foreground object with the bounding box represented by the coordinates of its top-left corner and bottom-right corner $(x_0, y_0, x_1, y_1)$, we create a rectangular cropping window with coordinates $(\frac{x_0}{2}, \frac{y_0}{2}, \frac{x_1+W}{2},\frac{y_1+H}{2} )$, where $W$ and $H$ are the width and height of the original image. The cropped patch is then resized to the input image size while keeping its aspect ratio. 

When the foreground object is relatively large in size, i.e.,  $ \mu > \pi_h$, the support image is downsized to create its augmented version. As illustrated in Figure \ref{fig1}(b), we downsize the image using bilinear interpolation with a resizing factor of $0.7$. The resulting image is then padded with gray pixel values to the input image size. 

On the other hand, if  $ \pi_l < \mu < \pi_h$, IDA is not triggered,  and only the original support image is included in the final support set.

\begin{figure*}
\begin{center}
\includegraphics[width=1.0\linewidth]{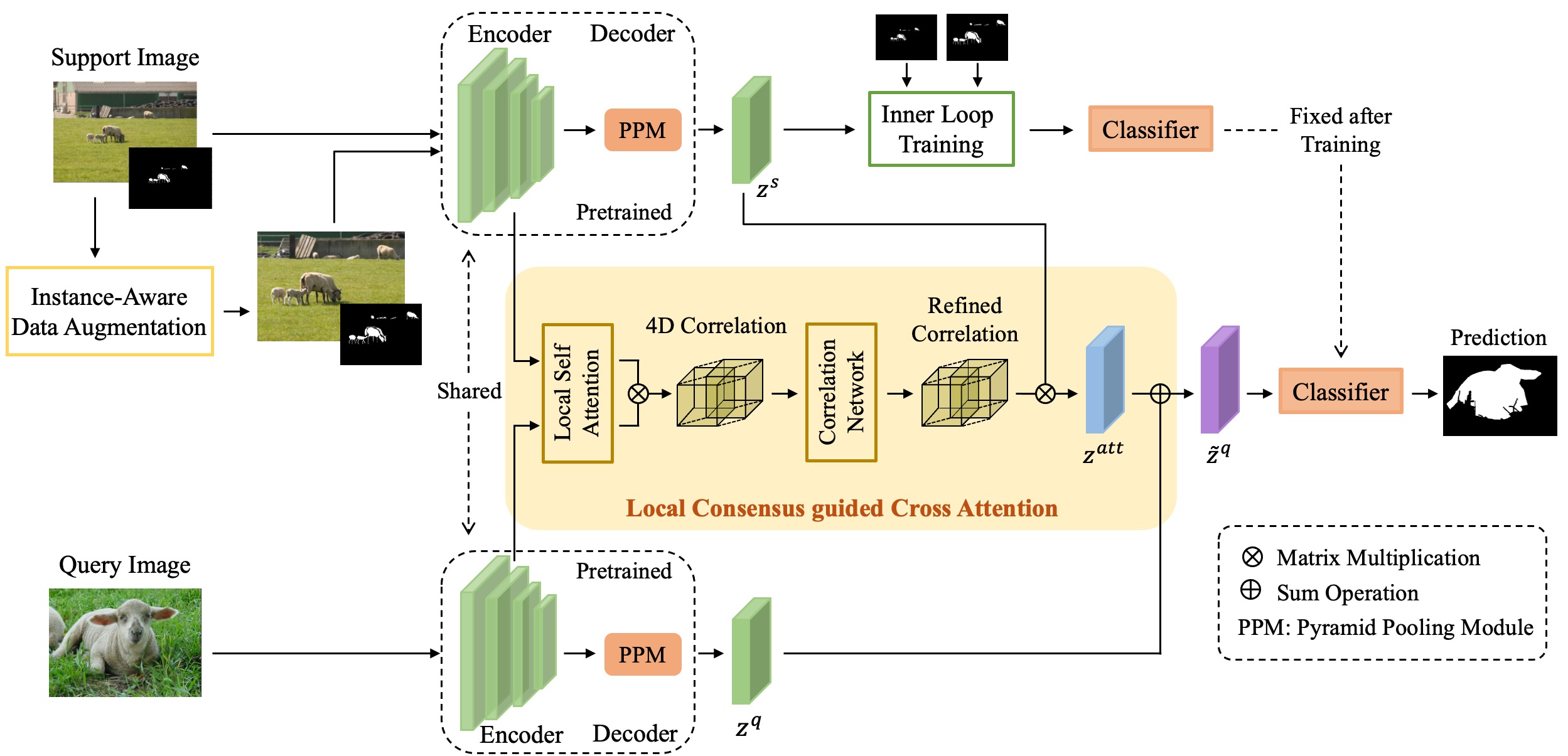}
\end{center}
   \caption{Overview of  LC-CAN,  which is trained in two stages. In the first stage, the backbone encoder and decoder are pre-trained on base classes. In the second stage, we meta-learn the LCCA module in an episodic manner. 
   At inference time, we train the classifier with the IDA-augmented support set and then pass the LCCA-aligned query feature to the learned classifier for query mask prediction. Note that LCCA module is based on features from multiple intermediate layers and the diagram only illustrates one layer for simplicity.}
   
\label{fig:lcca}
\end{figure*}

\begin{figure}
\begin{center}
\includegraphics[width=0.9\linewidth]{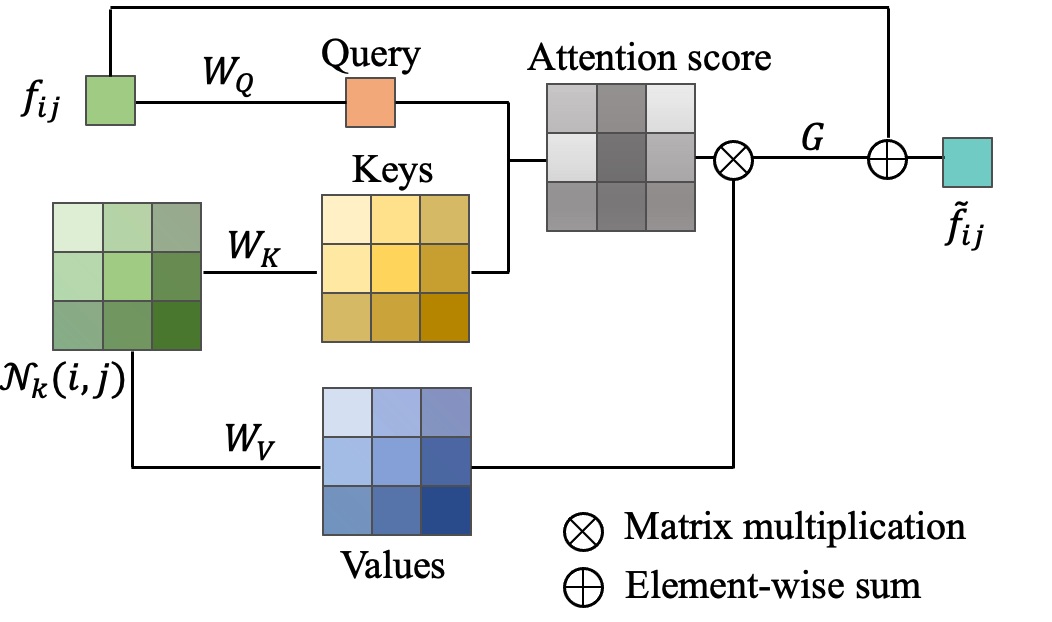}
\end{center}
   \caption{The local self-attention layer with spatial extent of $k=3$.}
   
\label{fig:lsa}
\end{figure}

\textbf{Why not use random augmentation?} 
Random data augmentation has proved effective in regular semantic segmentation tasks. However, only a few support images are available under the setting of FSS. Therefore, augmenting the support set using random data augmentation can drastically change the overall distribution of the support set and may create a bigger distribution discrepancy between support and query images, which will hinder the model's generalization ability to the query image. 
On the other hand,  IDA creates the augmented image adaptively based on the relative size of the target objects. It not only improves the support set's diversity but also helps correct the distribution bias of the support images. 

\subsection{Local Consensus Guided Cross Attention} \label{ssec: lcca}

When there are significant visual differences between support and query images, the feature distinction between the image pair may hinder the knowledge transfer and cripple the segmentation performance. Therefore, we propose to use cross-attention to align the query feature with the support feature and reduce the feature distinction. 
To formulate the Local Consensus guided Cross Attention (LCCA), we first establish some notations. Given a pair of support and query images, $x^s, x^q \in \mathbb{R}^{3 \times H \times W}$, we use the pre-trained backbone encoder to generate intermediate feature maps:


\begin{equation}\label{eq:1}
    f^{s}_1, \cdots, f^s_L = E_{\phi}(x^s) ,
\end{equation}
\begin{equation}\label{eq:2}
    f^{q}_1, \cdots, f^q_L = E_{\phi}(x^q), 
\end{equation}
with $f^{\sqcdot}_l \in \mathcal{R}^{c_l \times h_l \times w_l} (l=1, \cdots , L)$ being the intermediate output of the $l^{\text{th}}$ encoder block. 
And we use the pre-trained backbone decoder to extract the final feature embeddings:
\begin{equation}\label{eq:3}
    z^{s} = D_{\psi}(f^s_L) \in \mathbb{R}^{c \times h \times w} ,
\end{equation}
\begin{equation}\label{eq:4}
    z^{q} = D_{\psi}(f^q_L) \in \mathbb{R}^{c \times h \times w} .
\end{equation}

The feature embedding $z^{\sqcdot}$ is directly passed to a binary classifier $\hat{m}^{}=p_{\theta}(z^{\sqcdot})$ to separate foreground and background pixels. The classifier $p_{\theta}$ is trained with standard cross-entropy loss on the support set. 

Under few-shot settings, the learned classifier $p_{\theta}$ tends to overfit support images. To improve the model's generalizability to the query image, our proposed Local Consensus guided Cross Attention Network (LC-CAN) aligns the query feature $z^q$ with support feature $z^s$ and makes the final prediction based on the aligned query feature. Its overall diagram is illustrated in Figure \ref{fig:lcca}. The Local Consensus Guided Cross Attention (LCCA) is the key component in LC-CAN to perform feature alignment between support and query images using their dense correlation. Particularly, the proposed LCCA module consists of the following three components.



\textbf{Local Self-Attention.} 
Before constructing the cross-affinity between the support and query image pair, we first enhance their feature representation using local self-attention to capture more reliable local contextual information. 

Similar to conventional transformers \cite{vaswani2017attention, zhang2020feature}, local self-attention operates on {\it{queries}}, {\it{keys}}, and {\it{values}} of an input feature map $f\in \mathbb{R}^{c \times h \times w}$ and outputs a transformed feature map $\tilde{f}$ of the same shape.   
And unlike the global attention, we only compute self-attention within the local window of each pixel. As illustrated in Figure \ref{fig:lsa}, given a pixel $f_{ij} \in \mathbb{R}^c$,  we first extract its $k \times k$ neighborhood region $\mathcal{N}_k(i,j)$ which is centered around $f_{ij}$.
We pass pixel $f_{ij}$ and its neighborhood pixels $f_{ab} \bigl(a,b\in \mathcal{N}_k(i,j) \bigr)$ to linear transformations to get the query  $q_{ij}=W_Q f_{ij}$, keys  $k_{ab}=W_K f_{ab}$, and values  $v_{ab}=W_V f_{ab}$. The local self-attention computes the output at location $ij$ as: 
\begin{equation} 
    \tilde{f}_{ij} = f_{ij} +  {G} \Bigl( \smashoperator{\sum_{\mkern50mu a,b\in\mathcal{N}_k(i,j)}} \text{softmax}_{ab}(q_{ij}^T k_{ab})v_{ab} \Bigl)
\end{equation}
where  $\text{softmax}_{ab}$ denotes the softmax computed across all pixels in neighborhood $\mathcal{N}_k(i, j)$ and ${G}$ is a transformation function implemented with $1 \times 1$ convolution. We set $k$ to $3$ in our experiments, meaning each pixel only attends to the pixels in its $3\times 3$ neighborhood area.

\textbf{Correlation Network.} 
We then build the correlation map between support and query images based on the refined feature maps. Particularly, a pair of intermediate features  $f_l^q$ and $f_l^s$ is first passed to the local self-attention to get refined features $\tilde{f^q_l}$ and  $\tilde{f^s_l}$. Then we compute the pixel-wise cosine similarities between $\tilde{f^q_l}$ and  $\tilde{f^s_l}$, and obtain a $4$D correlation map $\mathcal{C}_l \in \mathbb{R}^{h_l\times w_l \times h_l \times w_l}$ with
\begin{equation}
    \mathcal{C}_l(i,j,a,b) = \frac{\langle \tilde{f}^q_l(i,j), \tilde{f}^s_l(a,b) \rangle} {\lVert \tilde{f}^q_l(i,j) \lVert \cdot \lVert \tilde{f}^s_l(a,b) \lVert}, 
\end{equation}
where $(i,j)$ and $(a,b)$ denote 2-dimensional spatial positions of the query and support features respectively. 
The correlation maps obtained from different layers $\{\mathcal{C}_l\}_{l=1}^L$ are then stacked together along the channel dimension after bi-linear interpolation to the size of $h \times w \times h \times w$, resulting in the final multi-channel correlation map $\mathcal{C} \in \mathbb{R}^{L \times h \times w \times h \times w}$. 

\begin{figure}
\begin{center}
\includegraphics[width=0.9\linewidth]{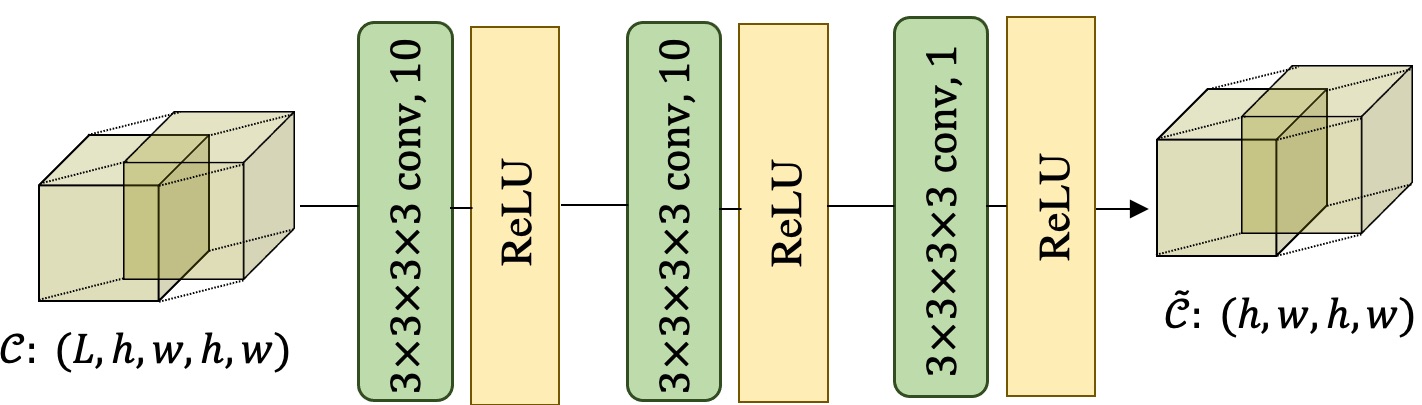}
\end{center}
   \caption{Structure of the correlation network, which is composed of a sequence of multi-channel 4D convolution units.}
   
\label{fig:corr}
\end{figure}

Due to the strong appearance difference between the image pair, great majority of the information in the correlation map corresponds to noisy matching. Inspired from \cite{rocco2018neighbourhood, zhao2021multi}, we adopt  the neighborhood consensus module to refine the correlation map:
\begin{equation}
    \tilde{\mathcal{C}} = H(\mathcal{C}) \in \mathbb{R}^{h \times w \times h \times w},
\end{equation} 
where the correlation network $H$ is composed of a sequence of multi-channel $4$D convolution units to refine the correlation map using local consensus constraints. The structure of $H$ is illustrated in Figure \ref{fig:corr} which is similar to \cite{rocco2018neighbourhood} except that our model takes a multi-channel correlation map as input to exploit diverse levels of feature representations. 
We refer the reader to \cite{rocco2018neighbourhood} for more details.
To reduce the computational burden caused by high-dimensional convolutions, we replace the original $4$D convolution with a lightweight center-pivot $4$D convolution \cite{min2021hypercorrelation}. 


\textbf{Attentive Feature Alignment}. 
With the refined affinity map between the image pair, we align the query feature with support feature using cross attention and compute the attention feature at position $ij$ as:  
\begin{equation}
    z_{ij}^{att} = \sum_{ab} \text{softmax}_{ab}(\tau \tilde{\mathcal{C}}_{ijab}) z_{ab}^s
\end{equation}
where $\tilde{\mathcal{C}}_{ijab}$ is the refined correlation score between query pixel at position $ij$ and support pixel at position $ab$,  and $z_{ab}^s$ is the support feature obtained from the decoder backbone. The temperature $\tau$ is a hyper-parameter to control the softness of the attention, and it is set to $10$ in our experiments.

The final aligned query feature is then computed as: 
\begin{equation}\label{eq:9}
    \tilde{z}^{q} = (1-\gamma)z^q +  \gamma z^{att},
\end{equation}
where  $\gamma$ is a hyper-parameter balancing the two terms and is set to 0.1 in all experiments. The aligned query feature $\tilde{z}^{q}$ is then passed to the classifier $p_{\theta}$ to get the final mask prediction $\hat{m}^q$.

To provide  direct supervision on LCCA, we pass the intermediate feature $z^{att}$ directly to the classifier $p_{\theta}$. The dice loss \cite{milletari2016v} between $p_{\theta}(z^{att})$ and the ground-truth query mask $m^q$ is used as the meta-training objective.


\subsection{K-Shot Setting}
The proposed LC-CAN can be easily extended to $K$-shot setting. Particularly, given $K$ support image-mask pairs $\mathcal{S} = \{(x^s_k, m^s_k)\}_{k=1}^K $ and a query image $x^q$, we first train the classifier $p_{\theta}$ based on the support set. 
We then align the query feature with $K$ support features through the proposed LCCA module to provide $K$ aligned query features, which are averaged along the pixel dimension to get the final query feature. This query feature is then passed to the classifier $p_{\theta}$  to obtain the final query mask prediction.


\subsection{Incorporating IDA and LC-CAN}
It's worth mentioning that the proposed LC-CAN, combined with IDA, is synergistic in boosting the model performance. Remarkably, by incorporating IDA,  the classifier $p_\theta$ is optimized based on the support set augmented by IDA. In the inference stage, we first align the query feature with the support feature and then make a prediction based on the aligned query feature. Instead of aligning the query feature with the original support image, we align it with the augmented version, which is more balanced in terms of the foreground proportion $\mu$. In Section \ref{sec:ablation}, we conduct extensive experiments to verify the benefit of incorporating IDA and LC-CAN.

\section{Experiments}

We conduct extensive experiments in this section to validate the effectiveness of the proposed LC-CAN model.

\subsection{Datasets}

Following previous works \cite{shaban2017one,wang2019panet,boudiaf2021few}, we conduct extensive experiments on  two widely-used FSS benchmarks, namely PASCAL-$5^i$ \cite{shaban2017one} and COCO-$20^i$\cite{nguyen2019feature}.

PASCAL-$5^i$ is derived from PASCAL VOC 2012 \cite{everingham2010pascal} with SBD augmentation \cite{hariharan2011semantic}. It consists of 20 object categories that are evenly divided into four folds, with each fold containing five classes.
COCO-$20^i$ is created from MS-COCO \cite{lin2014microsoft} and includes mask-annotated images from 80 object classes that are divided into four folds, with each fold containing 20 classes. 
We conduct cross-validation over all the folds in the above two datasets. Specifically, for each fold $i$, we sample data from the other three folds to train the model. Then we randomly select episodic data from all images in fold $i$ and evaluate them with the trained model.

\subsection{Implementation Details} \label{ssec:implement}

\textbf{Pre-training.} Our model is implemented using PyTorch framework. We build our model based on PSPNet \cite{zhao2017pyramid} with ResNet-50 and ResNet-101 \cite{he2016deep} as backbones.  All backbones are initialized with ImageNet \cite{krizhevsky2017imagenet} pre-trained weights. Other layers are initialized by the default setting of PyTorch. We adopt the standard supervised learning to train the feature extractor (i.e., the encoder and decoder)  on each fold of the FSS dataset, which consists of 16/61 classes (including background) for PASCAL-$5^i$/COCO-$20^i$. We train the model for 100 epochs on  PASCAL-$5^i$ and 20 epochs for COCO-$20^i$ with cross-entropy loss as the objective function. To update the parameters, we use SGD optimizer with an initial learning rate of $2.5\mathrm{e}{-3}$ and cosine learning rate decay, and the momentum is set to $0.9$, and weight decay to $1\mathrm{e}{-4}$. We set the batch size to 12 and the input image size to 473. Label smoothing is used with the smoothing parameter $\epsilon=0.1$. As for data augmentations, we only use random mirror flipping.

\textbf{Episodic Training.}
After the pre-training stage, we adopt an episodic training procedure to meta-learn the LCCA module.  Note that five convolution blocks (e.g. conv1\_x $\sim$ conv5\_x) are included in the ResNet backbones. 
Our proposed LCCA module computes two layer-wise correlation maps from the output features of conv4\_x and conv5\_x, and then refines them to obtain the final matching score. The pre-trained encoder and decoder are kept frozen during the episodic training stage.

In this stage, we organize the training data of base classes $\mathcal{D}_{base}$ into episodes, each including a support set and query set from a randomly sampled class.
We conduct the meta-training in a two-loop manner \cite{snell2017prototypical}. The inner loop trains a classifier $p_{\theta}$ for the selected class for 100 iterations on the support set with cross-entropy supervision. SGD optimizer with learning rate $1\mathrm{e}{-1}$ is used. The outer loop trains the proposed LCCA with dice loss supervision on the query images. The outer loop is trained with SGD optimizer on PASCAL-$5^i$ for 5 epochs and COCO-$20^i$ for 1 epoch. The learning rate is set to  $1\mathrm{e}{-3}$ on both datasets.

\textbf{Evaluation Metrics.} For evaluation metrics, we adopt the widely used mean Intersection over Union (mIoU), which is computed by averaging the IoU values of all classes in a fold. The formal definition of mIoU is as follows:
\begin{equation}
    mIoU = \frac{1}{C} \sum_{i=1}^{C}IoU_{i} ,
\end{equation}
where $C$ is the number of classes in each fold and $IoU_{i}$ is the intersection-over-union value of class $i$.  Following previous works \cite{wang2019panet,lang2022dcp}, for each fold, the model is validated on 1000 randomly sampled episodes.

\setlength{\extrarowheight}{2.65pt}


\begin{table*}[t]
\begin{center}
  { 
   
\begin{tabular}{l l l | l l l l l | l l l l l  }
    \hline

    \multirow{2}{*}{Backbone} & \multirow{2}{*}{Method}  & \multirow{2}{*}{Publication}  & \multicolumn{5}{c|} {1-shot} &  \multicolumn{5}{c} {5-shot}\\[2pt]  
                                                     & & & Fold-0 & Fold-1 & Fold-2 & Fold-3 & Mean 
                                                        & Fold-0 & Fold-1 & Fold-2 & Fold-3 & Mean \\
    \hline 
    
    \multirow{5}{*}{ResNet-50} 
                               & CANet \cite{zhang2019canet} & CVPR19 & 52.5 & 65.9 & 51.3 & 51.9 & 55.4 & 55.5 & 67.8 & 51.9 & 53.2 & 57.1    \\
                               & PGNet\cite{zhang2019pyramid} &  ICCV19 &56.0 &66.9 &50.6 &50.4 &56.0  & 57.7 &68.7 &52.9 &54.6 &58.5 \\ 
                               & PMMs \cite{yang2020prototype} &  ECCV20 & 55.2 & 66.9 & 52.6 & 50.7 & 56.3 & 56.3 & 67.3 & 54.5 & 51.0 & 57.3   \\[2pt]
                               & PFENet \cite{tian2020prior} &  TPAMI20 & 61.7 & 69.5 & 55.4 & {56.3} & 60.8 & 63.1 & 70.7 & 55.8 & 57.9 & 61.9 \\
                               & RePRI \cite{boudiaf2021few} & CVPR21 & 60.2 & 67.0 & {61.7} & 47.5 & 59.1 & 64.5 & 70.8 &  {71.7} & 60.3 & 66.8 \\
                               & HSNet \cite{min2021hypercorrelation} &  CVPR21 & {64.3} & {70.7} & 60.3 & \textbf{60.5} & \textbf{64.0} & \textbf{70.3} & {73.2} & 67.4 & \textbf{67.1} & {69.5} \\
                               & DCP \cite{lang2022dcp} &  IJCAI22 &  {63.8} &  {70.5} & {61.2} & 55.7 &  {62.8} &  {67.2} & \textbf{73.2} & 66.4 &  {64.5} & 67.8 \\
                               & MFNet \cite{zhang2022mfnet} &  TCSVT22 & \quad- & \quad- & \quad- & \quad- & 60.9 & \quad- & \quad- & \quad- & \quad- & 62.3 \\
                               &RPMG \cite{zhang2023rpmg} &  TCSVT23 & \textbf{64.4} & \textbf{72.6} & 57.9 & 58.4 & 63.3 & 65.3 & 72.8 & 58.4 & 59.8 & 64.1 \\

                               \cline{2-13}
                               & Baseline & & 55.0 & 62.5 & 60.6 & 47.5 & 56.4 & 61.5 & 70.7 & 72.5 & 59.7 & 66.1 \\
                               & {LC-CAN } & & 60.0 & 65.0 & {61.7} & 52.8 & 59.9  & 67.1 & {72.8} & {74.3} & 64.0 & {69.6} \\
                               & LC-CAN+RePRI & & 61.6 & 65.8 & \textbf{62.4} & 52.8 & 60.7   &  68.0 & \textbf{73.9} & \textbf{74.9} & 64.3 & \textbf{70.3} \\
                               \hline
                               
    \multirow{5}{*}{ResNet-101}
                               & PFENet \cite{tian2020prior} & TPAMI20 & 60.5 & {69.4} & 54.4 & {55.9} & 60.1 & 62.8 & 70.4 & 54.9 & 57.6 & 61.4 \\
                               & CWT \cite{lu2021simpler} & CVPR21 & 56.9 & 65.2 & 61.2 & 48.8 & 58.0  & 62.6 & 70.2 & {68.8} & 57.2 & 64.7 \\
                               & RePRI \cite{boudiaf2021few} &	CVPR21 & 59.6 & 68.6 & \textbf{62.2} & 47.2 & 59.4 & 66.2 & 71.4 & 67.0 & 57.7 & 65.6 \\
                               & HSNet \cite{min2021hypercorrelation} & CVPR21 & \textbf{67.3} & \textbf{72.3} & {62.0} & \textbf{63.1} & \textbf{66.2} & \textbf{71.8} & \textbf{74.4} & 67.0 & \textbf{68.3} & {70.4} \\
                               & RPMG \cite{zhang2023rpmg} & TCSVT23 & 63.0 & 73.3 & 56.8 & 57.2 & 62.6 & 67.1 & 73.3 & 59.8 & 62.7 & 65.7 \\
                               
                               \cline{2-13}
                               & Baseline & & 56.3 & 63.1 & 59.2 & 47.7 & 56.6  & 63.1 & 70.5 & 70.0 & 59.0 & 65.7 \\
                               & LC-CAN  & & {62.3} & 67.3 & 61.2 & 53.1 & {61.0} & {69.3} & {73.5} & {72.8} & {64.7} & {70.1} \\
                               & LC-CAN+RePRI & & 63.2 & 67.6 & 61.8 & 53.5 & 61.5 &  70.1 & 73.8 & \textbf{73.6} & 65.1 & \textbf{70.7} \\

   \hline
\end{tabular}

}
\end{center}
\caption{Comparison with state-of-the-art methods on PASCAL-$5^i$ in terms of mIoU. The $1\ts{st}$  method is \textbf{bold} .}
\label{table1}
\end{table*}

\begin{table*}[t]
  \begin{center}
    {\resizebox{0.98\textwidth}{!}{
\begin{tabular}{l l l | l l l l l | l l l l l  }
    \hline
    \multirow{2}{*}{Backbone} & \multirow{2}{*}{Method} & \multirow{2}{*}{Publication} & \multicolumn{5}{c|} {1-shot} &  \multicolumn{5}{c} {5-shot}\\ 
                                                      & & & Fold-0 & Fold-1 & Fold-2 & Fold-3 & Mean 
                                                        & Fold-0 & Fold-1 & Fold-2 & Fold-3 & Mean \\
    \hline 
    
    \multirow{5}{*}{ResNet-50} 
 
                               & PMMs \cite{yang2020prototype} & ECCV20 & 29.5 & 36.8 & 29.0 & 27.0 & 30.6  & 33.8 & 42.0 & 33.0 & 33.3 & 35.5 \\ 
                               
                               & CWT \cite{lu2021simpler} & CVPR21 & 32.2 & 36.0 & 31.6 & 31.6 & 32.9  & 40.1 & 43.8 & 39.0 & 42.4 & 41.3 \\
                               
                               & RePRI \cite{boudiaf2021few} & CVPR21 & 31.2 & 38.1 & 33.3 & 33.0 & 34.0 & 38.5 & 46.2 & 40.0 & 43.6 & 42.1 \\
                               
                               & HSNet \cite{min2021hypercorrelation} & CVPR21 & {36.3} & {43.1} & {38.7} & \textbf{38.7} & {39.2} & 43.3 & \textbf{51.3} & \textbf{48.2} & 45.0 & {46.9} \\ 

                               & DCP \cite{lang2022dcp} & IJCAI22 & \textbf{40.9} & \textbf{43.8} & \textbf{42.6} & {38.3} & \textbf{41.4} & {45.8} & 49.7 & 43.7 & {46.6} & 46.5 \\

                               & MFNet \cite{zhang2022mfnet} & TCSVT22 &  \quad - &  \quad - &  \quad - &   \quad - & 34.9 &\quad - & \quad - & \quad - & \quad - & 39.2 \\

                               & RPMG \cite{zhang2023rpmg} & TCSVT23 & 38.3 & 41.4 & 39.6 &  35.9 & 38.8 &\quad - & \quad - & \quad - & \quad - & \quad - \\

                               \cline{2-13}
                               & Baseline & & 30.5 & 34.8 & 30.6 & 33.2 & 32.3 & 42.6 & 45.3 & 40.4 & 43.1 & 42.9 \\
                               & LC-CAN  &  & 35.2 & 39.6 & 37.2 & {38.3} & 37.6 & {46.0} & {50.8} & {45.4} & {48.1} & {47.6} \\
                               & LC-CAN+RePRI & & 35.8 & 40.8 & 36.8 & 38.6 & 38.0 & \textbf{46.9} &51.1 &45.3 & \textbf{48.8} & \textbf{48.0} \\

                               \hline
                               
    \multirow{5}{*}{ResNet-101}
                               & CWT \cite{lu2021simpler} & CVPR21 & 30.3 & 36.6 & 30.5 & 32.2 & 32.4  & 38.5 & 46.7 & 39.4 & 43.2 & 42.0 \\
                               
                               & PFENet \cite{tian2020prior} & TPAMI20 & 36.8 & 41.8 & 38.7 & 36.7 & 38.5 & 40.4 & 46.8 & 43.2 & 40.5 & 42.7 \\ 
                               & HSNet \cite{min2021hypercorrelation} & CVPR21 & {37.2} & \textbf{44.1} & \textbf{42.4} & \textbf{41.3} & \textbf{41.2} & {45.9} & {53.0} & \textbf{51.8} & {47.1} & {49.5} \\ 
                               \cline{2-13}
                               & Baseline & & 33.0 & 38.7 & 32.2 & 34.7 & 34.7 & 42.2 & 49.3 & 43.7 & 43.6 & 44.7 \\
                               & LC-CAN &  & {37.7} & {42.7} & {40.0} & {39.8} & {40.1} &  {47.1} &  {54.4} & {48.6} &  {50.1} &  {50.0} \\
                               & LC-CAN+RePRI & & \textbf{38.5} & 43.6 & 39.5 & 40.1 & 40.4 & \textbf{47.7} & \textbf{55.0} & 48.0 & \textbf{50.1} & \textbf{50.2} \\
                               
   \hline
\end{tabular}
}}
\end{center}
\caption{
Comparison with state-of-the-art methods on COCO-$20^i$ in terms of mIoU.  The $1\ts{st}$  method is \textbf{bold} . }
\label{table2}
\end{table*}

\subsection{Experiment results}

In Table \ref{table1} and \ref{table2}, we evaluate LC-CAN under standard 1-shot and 5-shot settings on PASCAL-$5^i$ and COCO-$20^i$ datasets. Our final implementation of LC-CAN has integrated IDA for support set augmentation. We present the performance of LA-CAN along with the two-stage fine-tuning baseline which is introduced in Section \ref{ssec:overall}. 
Our proposed method shows a significant performance gain over the baseline, which verifies the effectiveness of the proposed approach. Specifically, our segmentation model achieves comparable results to recent prototype learning based approaches under 1-shot setting, and outperforms existing methods by a sizable margin under the 5-shot setting.

The COCO-$20^i$ dataset is considered more challenging than PASCAL-$5^i$ as it features a larger number of object categories and more substantial scale variations between objects.  The proposed LC-CAN framework effectively addresses these challenges. Particularly, LC-CAN effectively reduces the scale difference with IDA data augmentation and alleviates the visual disparity between support and query images using LCCA. Our experiments demonstrate that LC-CAN brings more significant improvements in segmentation accuracy on the COCO-$20^i$ dataset.

Furthermore, the LC-CAN framework is versatile and can be combined with other fine-tuning based approaches. In particular, we combined LC-CAN with RePRI \cite{boudiaf2021few} by adding an information maximization loss \cite{boudiaf2020information, boudiaf2021few}  as a regularization term in the process of classifier fine-tuning, which we refer to \texttt{LC-CAN+RePRI}. As illustrated in Table \ref{table1} and \ref{table2}, this resulted in further improvements in segmentation accuracy on both PASCAL-$5^i$ and COCO-$20^i$ datasets.

\subsection{Robustness to Domain Shift} 
We also evaluate our approach under a more challenging and realistic cross-domain setting and test the model's generalizability to unseen domains with different data distributions. Notably, we train the few-shot segmentation model on each fold of COCO-$20^i$ and then directly test the model on the novel tasks sampled from the PASCAL-$5^i$ dataset after removing the classes seen during the training phase. The results of the 1-shot and 5-shot cross-domain experiments are summarized in Table \ref{table:domain}. 
We observe that our model performs robustly in the presence of a large domain shift. Especially, it surpasses RePRI \cite{boudiaf2021few} by $3.4\%$ under the 5-shot setting.

\begin{table} 
  \begin{center}
    \resizebox{0.8\columnwidth}{!}{
\begin{tabular}{l l l l}
    \hline
    \multirow{2}{*}{Method} & \multirow{2}{*}{Backbone} & \multicolumn{2}{c} {COCO $ \rightarrow $ PASCAL} \\
    && 1 shot & 5 shot \\
    \hline 
    PFENet \cite{tian2020prior}  &  ResNet50  &  61.1 & 63.4 \\
    RePRI  \cite{boudiaf2021few} &  ResNet50  &  \textbf{63.2} & 67.4 \\
    CWT \cite{lu2021simpler}    &  ResNet50  & 59.6 & 66.5 \\
    LC-CAN(ours) &  ResNet50  & 61.7 & \textbf{70.8} \\
    \hline
\end{tabular}
}
\end{center}
\caption{Domain-shift results averaged over 4 folds, on COCO-$20^i$ to PASCAL. Best results are in bold.}
\label{table:domain}
\end{table}




\section{Ablation study}\label{sec:ablation}

In this section, we conduct extensive ablation studies to investigate the impact of major components in our model.  The experiments in this section are performed on COCO-$20^i$ dataset using the ResNet-50 backbone unless specified otherwise. 

\setlength{\extrarowheight}{2.7pt}

\begin{table}
  \begin{center}
    {\resizebox{0.97\columnwidth}{!}{
\begin{tabular}{l| l l l l  l}
    \hline
    \multirow{2}{*}{Variants} & \multicolumn{5}{c} {1-shot mIoU} \\
      & Fold-0 & Fold-1 & Fold-2 & Fold-3 & Mean \\
    \hline 
    Baseline      & 30.5 & 34.8 & 30.6 & 33.2 & 32.3 \\
    Baseline+RDA   & 31.3 & 34.6  & 29.8 & 32.3  & 32.0 \\ 
    Baseline+IDA  & \textbf{32.5} & \textbf{35.7}  & \textbf{31.9} & \textbf{33.6}  & \textbf{33.4} \\
    \hline
\end{tabular}
}}
\end{center}
\caption{Ablation studies of IDA under 1-shot setting. ‘RDA’ refers to the  naive random  data augmentation. }
\label{table:ida}
\end{table}

\begin{table}
  \begin{center}
    \resizebox{1\columnwidth}{!}{
\begin{tabular}{l| l l l l  l}
    \hline
    \multirow{2}{*}{Variants} & \multicolumn{5}{c} {1-shot mIoU} \\
      & Fold-0 & Fold-1 & Fold-2 & Fold-3 & Mean \\
    \hline 
    Baseline  & 30.5 & 34.8 & 30.6 & 33.2 & 32.3 \\
    IDA $[0.1, 0.2]$ & 32.4 & 36.1  & 31.0 & 33.1  & 33.2 \\ 
    IDA $[0, 0.3]$    & 31.4 & 34.9  & 30.9 & 33.7 & 32.7 \\ 
    IDA $[0.15, 1]$    & 31.5 &  35.2 & 31.6 & 34.1 & 33.1 \\ 
    \rowcolor{Gray} 
    IDA $[0.15, 0.3]$ & 32.5 & 35.7  & 31.9 & 33.6  & 33.4 \\

    \hline
\end{tabular}
}
\end{center}
\caption{Ablation studies of IDA on COCO-$20^i$ under 1-shot setting. In the table, IDA with thresholds $\pi_l$ and $\pi_h$ is represented as IDA $[\pi_l, \pi_h]$.  The final adopted setting is highlighted in gray. }
\label{table:ida1}
\end{table}

\subsection{Ablation on IDA}
To inspect the impact of the proposed IDA, we compare the performance of the following three models: (1) the \texttt{Baseline} model, which employs a two-stage fine-tuning approach, as introduced in Section \ref{ssec:overall}, (2) the \texttt{Baseline+IDA} model, which uses IDA to augment the support set during the testing stage and trains the classifier based on the augmented support set, and (3) the \texttt{Baseline+RDA} model, which adopts a naive random augmentation in the classifier fine-tuning stage. Specifically, we randomly resize and crop an image with a resize range of 0.5 to 1.5 times its original size.

 As shown in Table \ref{table:ida}, our proposed IDA strategy brings $1.1\%$ improvement in mIoU compared to the baseline model. Random augmentation may hurt the model's performance on the query set. As explained in Section \ref{ssec:ida}, the potential reason is that the IDA augmentation strategy applies augmentation adaptively based on the target object's relative size and helps correct the distribution bias in the support set. In contrast, random augmentation may create a bigger distribution discrepancy between the support and query sets.

Furthermore, we conducted an analysis of the hyperparameters $\pi_l$ and $\pi_h$ to investigate their impact on the IDA strategy's performance. Table \ref{table:ida1} shows that the default configuration IDA $[0.15, 0.3]$ with $\pi_l=0.15$ and $\pi_h=0.3$ outperforms other configurations. However, IDA $[0.1, 0.2]$ yields comparable performance to the default configuration, which indicates that the IDA strategy's performance is relatively robust to the choice of hyperparameters.
Notebly, when $\pi_l$ was set to $0$, IDA only performed image downsizing for support images with large foreground objects. On the other hand, when $\pi_h$ was set to $1.0$, IDA only performed instance-aware crop for support images with a small foreground ratio $\mu$. Both image downsizing and instance-aware crop improved the model's performance over the baseline method. Ultimately, the complete implementation of IDA, which integrated both scenarios, achieved a performance gain of $1.1\%$ over the baseline.


The proposed IDA is a general solution for adaptive data augmentation under the few-shot settings, which can be seamlessly integrated into other FSS approaches to improve segmentation performance. In this study, we incorporate IDA into two prior approaches, including PFENet \cite{tian2020prior} and RePRI \cite{boudiaf2021few}, to verify the effectiveness of IDA. The experimental results on the COCO dataset (see Table \ref{table:ida_other}) demonstrate that IDA can consistently achieve higher performance for both prototype learning based \cite{tian2020prior} and fine-tuning based \cite{boudiaf2021few} approaches. Notably, our approach benefits greatly from the synergistic effects between LC-CAN and IDA, which lead to a further 2.1\% improvement in segmentation accuracy.

\setlength{\extrarowheight}{2.7pt}

\begin{table*} 
  \begin{center}
    \resizebox{0.73\textwidth}{!}{
\begin{tabular}{l| c | l l l l  l}
    \hline
    \multirow{2}{*}{Method} & \multirow{2}{*}{w / IDA} & \multicolumn{5}{c} {1-shot mIoU} \\
    &  & Fold-0 & Fold-1 & Fold-2 & Fold-3 & Mean \\
    \hline 
    \multirow{2}{*}{PFENet \cite{tian2020prior}} &     \xmark  & 36.5 & 38.6 & 34.5 & 33.8 & 35.8\\
    &  \checkmark  & 36.8 \textcolor{red}{(+0.3)} & 39.4 \textcolor{red}{(+0.8)} & 35.0 \textcolor{red}{(+0.5)} & 34.4 \textcolor{red}{(+0.6)} & 36.4 \textcolor{red}{(+0.6)} \\
    
    \hline 
    \multirow{2}{*}{RePRI \cite{boudiaf2021few} } &   \xmark  & 32.3 & 37.6 & 32.0 & 33.7 & 33.9\\
    &  \checkmark  & 33.0 \textcolor{red}{(+0.6)} & 38.1 \textcolor{red}{(+0.5)} & 33.2 \textcolor{red}{(+1.2)} & 34.8 \textcolor{red}{(+1.1)} & 34.8 \textcolor{red}{(+0.9)} \\

    \hline 
    \multirow{2}{*}{LC-CAN(Ours)} & \xmark & 33.4 & 37.3 & 34.5 & 36.6 & 35.5\\
    &  \checkmark  & 35.2 \textcolor{red}{(+1.8)} & 39.6 \textcolor{red}{(+2.3)} & 37.2 \textcolor{red}{(+2.7)} & 38.3 \textcolor{red}{(+1.7)} & 37.6 \textcolor{red}{(+2.1)} \\
    
    \hline
\end{tabular}
}
 \end{center}
\caption{Effectiveness of IDA with different approaches tested on COCO-$20^i$ under 1-shot setting. The experiment results are reproduced by us. The term w/IDA indicates whether the model uses IDA for support set augmentation.}
\label{table:ida_other}
\end{table*}

\begin{table}
  \begin{center}
   {\resizebox{0.97\columnwidth}{!}{
\begin{tabular}{l| l l l l  l}
    \hline
    \multirow{2}{*}{Variants} & \multicolumn{5}{c} {1-shot mIoU} \\
      & Fold-0 & Fold-1 & Fold-2 & Fold-3 & Mean \\
    \hline 
    Baseline       & 30.5 & 34.8 & 30.6 & 33.2 & 32.3\\
    LC-CAN         & \textbf{33.4} & \textbf{37.3} & \textbf{34.5} & \textbf{36.6} & \textbf{35.5} \\
    LC-CAN w/o LSA & 32.7 & 36.6 & 34.0 & 35.6 & 34.7 \\ 
    \hline
\end{tabular}
}}
\end{center}
\caption{Ablation studies of LC-CAN. ‘LSA’ denotes the local self-attention module in LCCA used for feature enhancement.}
\label{table:lcca}
\end{table}

\begin{table}
  \begin{center}
    \resizebox{1.0\columnwidth}{!}{
\begin{tabular}{l| l l l l  l}
    \hline
    \multirow{2}{*}{Variants} & \multicolumn{5}{c} {1-shot mIoU} \\
      & Fold-0 & Fold-1 & Fold-2 & Fold-3 & Mean \\
    \hline 
     Baseline  & 30.5 & 34.8 & 30.6 & 33.2 & 32.3 \\
     LC-CAN $\mathcal{P}^{5}$     & 32.5 & 36.3 & 33.8 & 35.9 & 34.6 \\ 
    \rowcolor{Gray} 
    LC-CAN $\mathcal{P}^{4:5}$    & 33.4 & 37.3 & 34.5 & 36.6 & 35.5 \\ 
     LC-CAN $\mathcal{P}^{3:5}$  & 32.5 & 37.2  & 33.7 & 36.1 & 34.9 \\ 
    \hline
\end{tabular}
}
\end{center}
\caption{Ablation studies on the pyramid layers of LCCA on COCO-$20^i$. The default setting adopted in the paper is highlighted in gray.}
\label{table:lcca1}
\end{table}

\subsection{Ablation on LCCA}
In this section, we investigate the effectiveness of LCCA by isolating its impact from the performance boost resulting from IDA. Specifically,  we train LC-CAN \textit{without} any data augmentation, and in the testing stage, we optimize the classifier $p_{\theta}$ based on the original support set \textit{without} IDA for data augmentation. Then with the learned classifier $p_{\theta}$, we predict the query mask based on the refined query feature, which is aligned with support images through LCCA. 

Table \ref{table:lcca} demonstrates that our proposed LC-CAN model significantly outperforms the fine-tuning baseline, indicating the effectiveness of the LCCA module in closing the feature discrepancy between support and query images. Additionally, removing the local self-attention component in LCCA, which enhances the features, results in a performance drop of $0.8\%$, as shown in the third row of Table \ref{table:lcca}. This finding highlights the importance of contextual information around the neighborhood pixels in refining the cross affinity. 

As introduced in Section \ref{ssec: lcca},  the LCCA module exploits correlation maps $\{\mathcal{C}_l\}_{l=1}^L$ obtained from multiple layers of the encoder backbone to obtain more accurate matching correspondence between support and query images. To investigate the impact of the correlation map $\mathcal{C}_l$ from each layer $l$, we trained and evaluated our model using different correlation pyramids: $\mathcal{P}^{3:5} = \{\mathcal{C}_3, \mathcal{C}_4, \mathcal{C}_5\}$, $\mathcal{P}^{4:5} = \{\mathcal{C}_4, \mathcal{C}_5\}$, and $\mathcal{P}^{5} = \{ \mathcal{C}_5 \}$, where $\mathcal{C}_l$ is the correlation map computed based on the output features of ResNet block $l$. To isolate impact of LCCA from the performance boost resulting from IDA,
these experiments were conducted {\it{without}} using IDA for support set augmentation, and the results are summarized in Table \ref{table:lcca1}. 
Our findings show that LCCA based on $\mathcal{C}^{4:5}$ achieves the best performance, indicating that the deep features from ResNet block 4 and 5 provide essential semantic information for improving the matching accuracy. In contrast, the shallow features related to low-level visual patterns may introduce noisy matching correspondence.

\begin{figure}[t]
\begin{center}
\includegraphics[width=1.03\columnwidth]{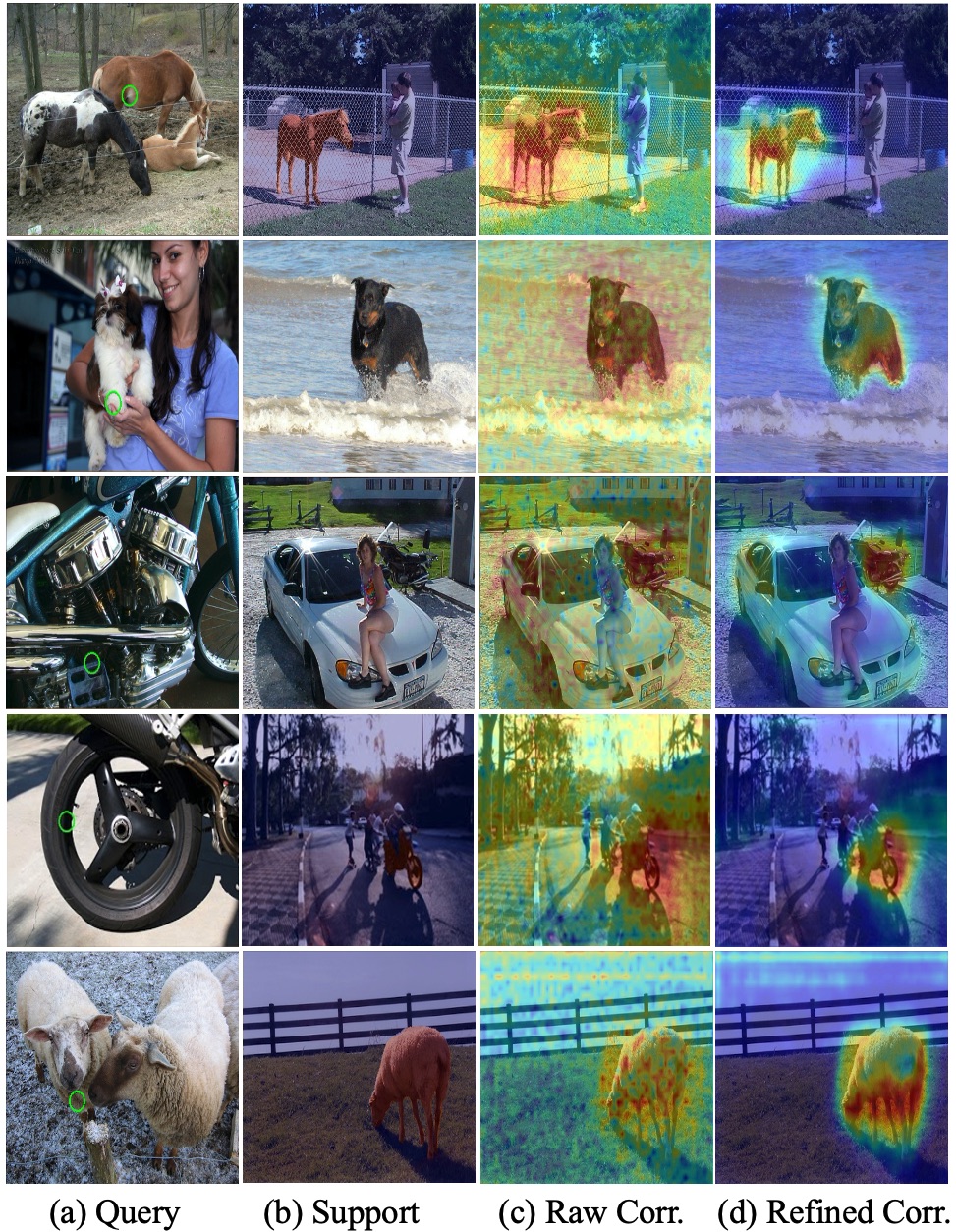}
\end{center}
   \caption{Visualization of the correlation maps between support and query images: (a) query image with the anchor pixel highlighted in green, (b) support image, (c) raw correlation map, (d) final correlation map refined by correlation network.}
\label{fig_corr}
\end{figure}

As introduced in Section \ref{ssec: lcca}, LCCA adopts a correlation network to refine the raw correlation map computed based on the feature map output from the encoder backbone. To verify the effectiveness of the correlation network in improving the matching accuracy, we visualize the raw correlation map computed from the output features of ResNet block 5 and the final correlation map refined by the correlation network. As illustrated in Figure \ref{fig_corr}, the raw correlation maps between support-query image pairs are very noisy due to the large visual difference. The correlation network effectively suppresses this noise and improves the accuracy of the correlation map. The refined correlation map is more coherent and accurate than the raw correlation map. Based on the refined correlation map, the LCCA module aligns the query feature with the support feature, leading to further improvements in segmentation performance on the query images.

\begin{table}
  \begin{center}
    {\resizebox{0.95\columnwidth}{!}{
\begin{tabular}{l| l l l l  l}
    \hline
    \multirow{2}{*}{Variants} & \multicolumn{5}{c} {1-shot mIoU} \\
      & Fold-0 & Fold-1 & Fold-2 & Fold-3 & Mean \\
    \hline 
    Baseline       & 30.5 & 34.8 & 30.6 & 33.2 & 32.3\\
    LC-CAN/both & 35.5 & 39.5 & 37.1 & 38.1 & 37.5 \\ 
    LC-CAN/org & 34.5 & 38.5 & 36.3 & 37.1 & 36.6\\
    \rowcolor{Gray}
    LC-CAN/aug  & 35.2 & 39.6 & 37.2 & 38.3 & 37.6 \\ 
    \hline
\end{tabular}
}}
\end{center}
\caption{Incorporating IDA and LC-CAN. The approach adopted in the final implementation is highlighted in gray.}
\label{table:inc}
\end{table}

\subsection{Incorporating IDA and LC-CAN} 
An intuitive way to incorporate IDA with LC-CAN is to align the query feature with both the original support image and its augmented version through LCCA to produce two aligned query features, the average of which is the final query embedding used for classification. We denote this method as \texttt{LC-CAN/both}.  However, this approach is costly regarding computation overhead and memory consumption. In Table \ref{table:inc}, we compare two other  strategies with the intuitive method mentioned above. The first strategy is to align the query feature with the original support image only, which is denoted as \texttt{LC-CAN/org}. The second strategy is to align the query feature with the augmented support image only, which is denoted as \texttt{LC-CAN/aug}. As shown in Table \ref{table:inc}, \texttt{LC-CAN/aug} achieves the best result despite its lower computation cost compared to \texttt{LC-CAN/both}. Aligning the query image with the augmented support image is more beneficial than the other two approaches since the augmented image is more balanced in foreground-background ratio. 

IDA augmentation strategy and the proposed LC-CAN framework cooperate in a synergistic fashion to improve the model's performance. For example, with a ResNet-50 backbone, IDA and LC-CAN bring $1.1\%$ and $3.2\%$ improvement on COCO-$20^i$ in 1-shot mIoU compared to the baseline. And incorporating both methods the final model achieves $5.3\%$ improvement over the baseline. 

\subsection{Qualitative Study}

\begin{figure*}
\begin{center}
\includegraphics[width=0.99\linewidth]{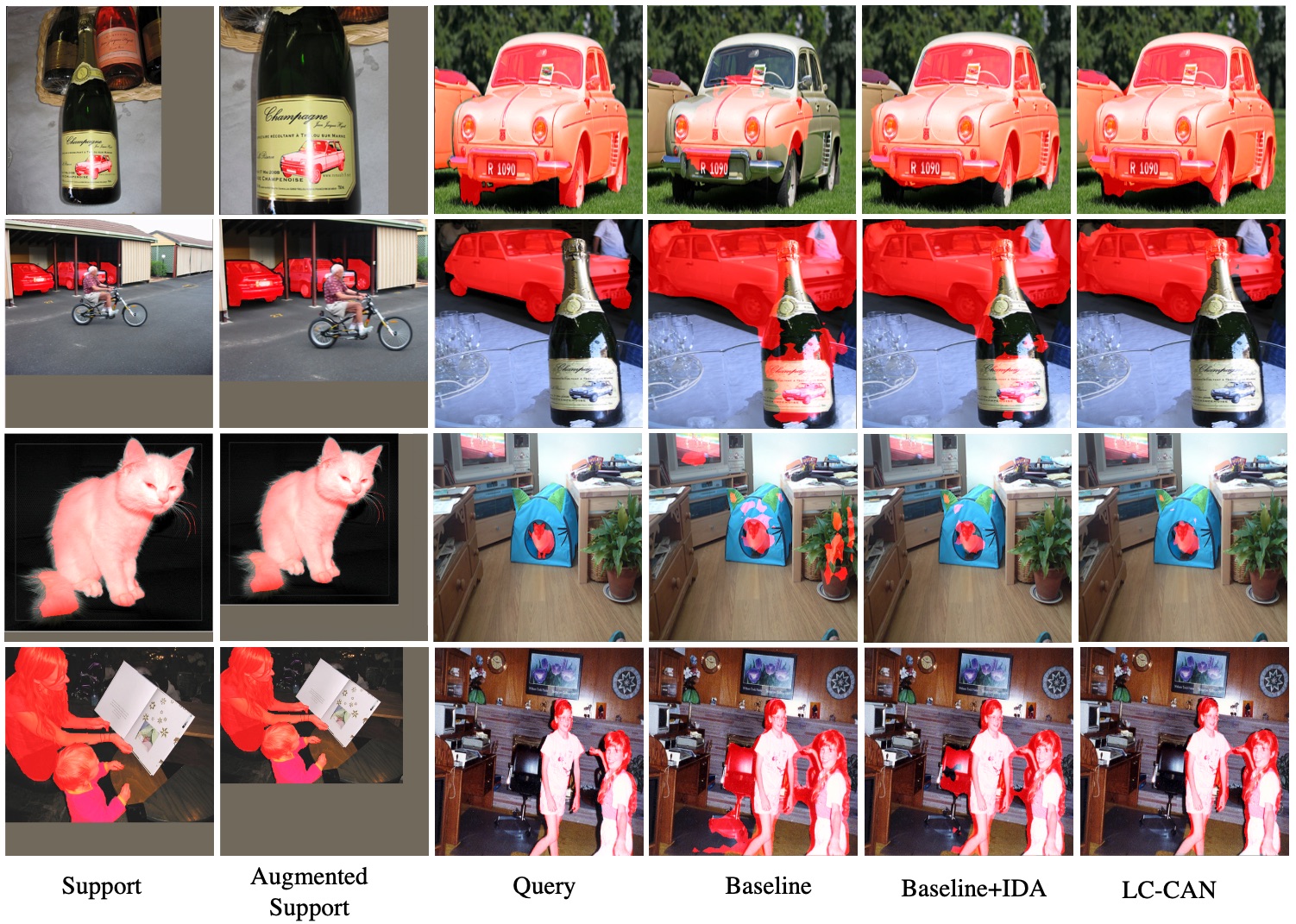}
\end{center}
   \caption{Case study for one-shot segmentation on PASCAL-5$^i$.  The case study covers two different scenarios: episodes with support images featuring extremely small target objects (the first two rows) and those with support images containing very large objects (the last two rows). 
   From left to right, in each column, we show (1) the original support image with the ground truth mask highlighted in red, (2) the support image augmented using IDA, (3) the query image with the ground truth mask outlined in red, (4) the segmentation mask prediction from the fine-tuning baseline model, (5) the segmentation mask prediction from improved fine-tuning strategy with IDA augmentation, and (6) the segmentation mask prediction from LC-CAN.}
\label{fig:case}
\end{figure*}

In Figure \ref{fig:case}, we analyze the effectiveness of the proposed LC-CAN qualitatively. Particularly, we compare the segmentation masks produced by the following three models: (1) the fine-tuning baseline model, (2) the improved fine-tuning strategy with IDA augmentation, and (3) the full implementation of LC-CAN with IDA. 

As shown in Figure \ref{fig:case}, when the support images contain extremely small or large scale objects, the IDA augmentation strategy effectively closes the distribution gap between the support and query images, which improves the model's generalizability to the query images. 
In addition,  when there are large visual differences between the support and query images, LC-CAN can take advantage of the dense correlation between support and query images to effectively suppress the noisy background area, which is falsely activated in the baseline model. 

\section{Conclusion}

Most recent literature on few-shot segmentation has adopted a prototype learning paradigm, which exhibits impressive performance under extreme low-shot settings. However, these approaches have certain limitations: (1) their performance saturates quickly beyond the standard 1- or 5-shot setting for semantic segmentation, and (2) they do not scale well to generalized few-shot segmentation or perform well in scenarios with domain shifts. In this paper, we re-evaluate fine-tuning based FSS approaches compared to the prototype learning paradigm. Specifically, we introduce an adaptive data augmentation strategy to promote distribution consistency between support and query images and alleviate the overfitting problem. Additionally, we incorporate the dense correlation between support and query images to improve the performance of fine-tuning based FSS approaches, particularly in extreme low-shot scenarios. By highlighting the relative advantages and disadvantages of fine-tuning based methods, we hope to inspire the development of new algorithms.



%

\bibliographystyle{IEEEtran}
\bibliography{egbib}

\end{document}